\def\year{2018}\relax
\documentclass[letterpaper]{article} 
\usepackage{aaai18}  
\usepackage{times}  
\usepackage{helvet}  
\usepackage{courier}  
\usepackage{url}  
\usepackage{graphicx}  
\usepackage{tabularx}
\usepackage{color}
\usepackage{makecell}
\newcommand{\rpm}{\raisebox{.2ex}
{$\scriptstyle\pm$}}
\begin{document}
%
\title{Explainable PCGML via Game Design Patterns}
\author{Matthew Guzdial$^1$, Joshua Reno$^1$, Jonathan Chen$^1$, Gillian Smith$^2$, and Mark Riedl$^1$\\
Georgia Institute of Technology$^1$\\
Worcester Polytechnic Institute$^2$\\
\{mguzdial3, jreno, jonathanchen\}@gatech.edu, gmsmith@wpi.edu, riedl@cc.gatech.edu\\
}
\maketitle
\begin{abstract}
Procedural content generation via Machine Learning (PCGML) is the umbrella term for approaches that generate content for games via machine learning.
One of the benefits of PCGML is that, unlike search or grammar-based PCG, it does not require hand authoring of initial content or rules.
Instead, PCGML relies on existing content and black box models, which can be difficult to tune or tweak without expert knowledge. 
This is especially problematic when a human designer needs to understand how to manipulate their data or models to achieve desired results. 
We present an approach to Explainable PCGML via Design Patterns in which the design patterns act as a vocabulary and mode of interaction between user and model.
We demonstrate that our technique outperforms non-explainable versions of our system in interactions with five expert designers, four of whom lack any machine learning expertise.

\end{abstract}

\section{Introduction}

Procedural Content Generation (PCG), represents a field of research into, and a set of techniques for, generating game content algorithmically. 
PCG historically requires a significant amount of human-authored knowledge to generate content, such as rules, heuristics, and individual components, creating a time and design expertise burden. Procedural Content Generation via Machine Learning (PCGML) attempts to solve these issues by applying machine learning to extract this design knowledge from existing corpora of game content \cite{summerville2017procedural}. However, this approach has its own weaknesses; Applied naively, these models require machine learning literacy to understand and debug. Machine learning literacy is uncommon, especially among those designers who might most benefit from PCGML. 

Explainable AI represents a field of research into opening up black box Artificial Intelligence and Machine Learning models to users \cite{biran2017explain}. The promise of explainable AI is not just that it will help users understand such models, but also tweak these models to their needs \cite{olah2018building}. If we could include some representation of an individual game designer's knowledge into a model, we could help designers without ML expertise better understand and alter these models to their needs.

Design patterns \cite{bjork2004pattern} represent one popular way to represent game design knowledge. A design pattern is a category of game structure that serves a general design purpose across similar games. Researchers tend to derive design patterns via subjective application of design expertise \cite{hullett2010fps}, which makes it difficult to broadly apply one set of patterns across different designers and games. The same subjective limitation also means that an individual set of design patterns can serve to clarify what elements of a game matter to an individual designer. Given a set of design patterns specialized to a particular designer one could leverage these design patterns in an Explainable PCGML system to help a designer understand and tweak a model to their needs. We note our usage of the term pattern differs from the literature. Typically, a design pattern generalizes across designers, whereas we apply it to indicate the unique structures across a game important to an individual designer.

\begin{figure*}[tb]
  \centering
  \includegraphics[height=1.25in]{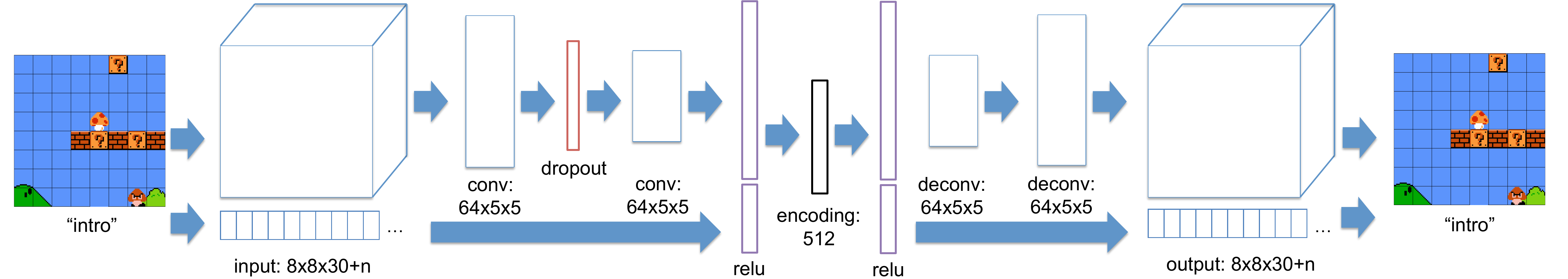}
  \caption{Network architecture for the Autoencoder.}
  \label{fig:autoencoder}
\end{figure*}

We present an underlying system for a potential co-creative PCGML tool, intended for designers without ML expertise. This system takes user-defined design patterns for a target level, and outputs a PCGML model. The design patterns provided by designers and generated by our system can be understood as labels on level structure, which allow our PCGML model to better represent and reflect the design values of an individual designer. This system has two major components: (1) a classification system that learns to classify level structures with the user-specified design pattern labels. This system ensures a user does not have to label all existing content to train (2) a level generation system that incorporates the user's level design patterns, and can use these patterns as a vocabulary with which to interact with the user. For example, generating labels on level structure to represent the model's interpretation of that structure to the user.

The rest of this paper is organized as follows. First, we relate our work to prior, related work. Second, we describe our Explainable PCGML (XPCGML) system in terms of the two major components. Third, we discuss the three evaluations we ran with five expert designers. We end with a discussion of the systems limitations, future work, and conclusions. Our major contributions are the first application of explainable AI to PCGML, the use of a random forest classifier to minimize user effort, and the results of our evaluations. Our results demonstrate both the promise of these pattern labels in improving user interaction and a positive impact on the underlying model's performance.

\section{Related Work}

There exist many prior approaches to co-creative or mixed-initiative design agents and editors \cite{yannakakis2014mixed,deterding2017mixed}. However, the majority of existing approaches have relied upon search or grammar-based approaches instead of machine learning, making it difficult to adapt to the needs of a particular designer over time \cite{liapis2013sentient,shaker2013ropossum,baldwin2017mixed}. A final version of our system would focus on machine learning, adapting to the user, and \textit{explaining and visualizing its inner model/process}.

Procedural content generation via Machine Learning \cite{summerville2017procedural} is a relatively new field, focused on generating content through machine learning methods. The majority of PCGML approaches represent black box methods, without any prior approach focused on explainability or co-creativity. We note some discussion in the Summerville et al. survey paper on potential collaborative approaches. Summerville \shortcite{summerville2016learning} explored adapting levels to players, but no work to our knowledge looks at adapting models to individual designers. 

Super Mario Bros. (SMB) represents a common area of research into PCGML \cite{dahlskog2012patterns,summerville2016super,jain2016autoencoders,snodgrass2017learning}. 
Beyond explainability, our approach differs from prior SMB PCGML approaches in terms of representation quality and the size of generated content. 
We focus on the generation of individual level sections instead of entire levels in order to better afford collaborative level building \cite{smith2011tanagara}. 
Second, prior approaches have abstracted away the possible level components into higher order groups. 
For example, treating all enemy types as equivalent and ignoring decorative elements. 
We make use of a rich representation of all possible level components and an ordering that allows our approach to place decorative elements appropriately.  

Explainable AI represents an emerging field of research \cite{biran2017explain}, focused on translating or rationalizing the behavior of black box models. To the best of our knowledge, this has not been previously applied to PCGML. 
Codella et al. \shortcite{codella2018teaching} demonstrated how explanations could improve model accuracy on three tasks, but required that every sample be hand-labeled with an explanation and treated explanations from different authors as equivalent.
Ehsan et al. \shortcite{ehsan2017rationalization} made use of explainable AI  for explainable agent behavior for automated game playing. Their approach relies on rationalization, which relies on a second machine learning interpretation of the original behavior, rather than visualizing or explaining the original model as our approach does. 

Design patterns represent a well-researched approach to game design \cite{bjork2004pattern}. In theory, game design patterns describe general solutions to game design problems that occur across many different games. Game Design patterns have been used as heuristics in evolutionary PCG systems including in the domain of Super Mario Bros. \cite{dahlskog2012patterns}. Researchers tend to derive game design patterns through either rigorous, cross-domain analysis \cite{milam2010design} or based upon their subjective interpretation of game structure. We embrace this subjectivity in our work by having designers create a language of game design patterns unique to themselves with which to interact with a PCGML system. 

\section{System Overview}

The approach presented in this paper builds an Explainable PCGML model based on existing level structure and an expert labeling design patterns upon that structure. We chose Super Mario Bros. as a domain given its familiarity to the game designers who took part in our evaluation. The general process for building a final model is as follows: First, users label existing game levels with the game design patterns they want to use for communicating with the system. For example, one might label both areas with large amounts of enemies and areas that require precise jumps as ``challenges''. The exact label can be anything as long as it is used consistently. Given this initial user labeling of level structure, we train a random forest classifier to classify additional level structure according to the labeled level chunks \cite{liaw2002classification}, which we then use to label all available levels with the user design pattern labels. Given this now larger training set of both level structure and labels, we train a convolutional neural network-based autoencoder on both levels structure and its associated labels \cite{lang1988time,lecun1989backpropagation}, which can then be used to generate new level structure and label its generated content with these design pattern labels \cite{jain2016autoencoders}.

We make use of Super Mario Bros. as our domain, and, in particular, we utilize those Super Mario Bros levels present in the Video Game Level Corpus \cite{summerville2016vglc}. We do not include underwater or boss/castle Super Mario Bros. levels. We made this choice as we perceived these two level types to be significantly different from all other level types. Further, while we make use of the VGLC levels, we do not make use of any of the VGLC Super Mario Bros. representations, which abstract away level components into higher order groups. Instead, we draw on the image parsing approach introduced in \cite{guzdial2016game}, using a spritesheet and OpenCV \cite{bradski2000opencv} to parse images of each level for a richer representation. 

In total we identified thirty unique classes of level components, and make use of a matrix representation for each level section of size $8 \times 8 \times 30$. The first two dimensions determine the tiles in the $x$ and $y$ axes, while the last dimension represents a one-hot vector of length 30 expressing component class. This vector is all 0's for any empty tile of a Super Mario Bros. level, and otherwise has 1's at the index associated with that particular level component. Thus, we can represent all level components, including background decoration. We note that we treat palette swaps of the same component as equivalent in class.

We make use of the SciPy random forest classifier \cite{jones2014scipy} and tensorflow for the autoencoder \cite{abadi2016tensorflow}. 

\subsection{Design Pattern Label Classifier}

Our goal for the design pattern label classifier is to minimize the amount of work and time costs for a potential user of the system. Users have to label level structure with the design patterns they would like to use, but the label classifier ensures they do not have to hand-label all available levels. The classifier for this task must be able to perform given access to whatever small amount of training data a designer is willing to label for it, along with being able to easily update its model given potential feedback from a user. We anticipate the exact amount of training data the system has access to will differ widely between users, but we do not wish to overburden authors with long data labeling tasks. Random forest classifiers are known to perform reasonably under these constraints \cite{michalski2013machine}.

The random forest model takes in an eight by eight level section and returns a level design pattern (either a user-defined design pattern or none). We train the random forest model based on the design pattern labels submitted by the user. 
We use a forest of size 100 with a maximum depth of 100 in order to encourage generality. 

In the case of an interactive, iterative system  the random forest can be easily retrained. In the case where the random forest classifier correctly classifies any  new design pattern there is no need for retraining. Otherwise, we can delete a subset of the trees of the random forest that incorrectly classified the design pattern, and retrain an appropriate number of trees to return to the maximum forest size on the existing labels and any additional new information.

Even with the design pattern level classifier this system requires the somewhat unusual step of labeling existing level structure with design patterns a user finds important. However, this is a necessary step for the benefit of a shared vocabulary, and labeling content is much easier than designing new content. Further, we note that when two humans collaborate they must negotiate a shared vocabulary.

\subsection{Generator}

The existing level generation system is based on an autoencoder, and we visualize its architecture in Figure \ref{fig:autoencoder}. The input comes in the form of a chunk of level content and the associated design patterns label, such as ``intro'' in the figure. This chunk is represented as an eight by eight by thirty input tensor plus a tensor of size $n$ where $n$ indicates the total number of design pattern labels given by the user. This last vector of size $n$ is a one-hot encoded vector of level design pattern labels.

After input, the level structure and design pattern label vector are separated. The level structure passes through a two layer convolutional neural network (CNN). We note that we placed a dropout layer in between the two CNN layers to allow better generalization. After the CNN layers the output of this section and the design patterns vector recombine and pass through a fully connected layer with relu activation to an embedded vector of size 512. We note that, while large, this is much smaller than the 1920+$n$ features of the input layer. The decoder section is an inverse of the encoder section of the architecture, starting with a relu fully connected layer, then deconvolutional neural network layers with upsampling handling the level structure. We implemented this model with an adam optimizer and mean square loss. Note that for the purposes of evaluation this is a standard autoencoder. We intend to make use of a variational autoencoder in future work \cite{kingma2013auto}.

\section{Evaluation}

Our system has two major parts: (1) a random forest classifier that attempts to label additional content with user-provided design patterns to learn the designer's vocabulary and (2) an autoencoder over level structure and associated patterns for generation. In this section we present three evaluations of our system. The first addresses the random forest classifier of labels, the second the entirety of the system, and the third addresses the limiting factor of time in human computer interactions. For all three evaluations we make use of a dataset of levels from Super Mario Bros. labeled by five expert designers. 

\subsection{Dataset Collection}

We reached out to ten design experts to label three or more Super Mario Bros. levels of their choice to serve as a dataset for this evaluation. We do not include prior, published academic patterns of Super Mario Bros. levels (e.g. \cite{dahlskog2012patterns}) as these patterns were designed for general automated design instead of explainable co-creation. Our goals for choosing these ten designers were to get as diverse a pool of labels as possible. Of these ten, five responded and took part in this study.

\begin{itemize}
  \item \textbf{Adam Le Doux}: Le Doux is a game developer and designer best known for his Bitsy game engine. He is currently a Narrative Tool Developer at Bungie.
  \item \textbf{Dee Del Rosario}: Del Rosario is an events and community organizer in games with organizations such as Different Games Collective and Seattle Indies, along with being a gamedev hobbyist. They currently work as a web developer and educator.
  \item \textbf{Kartik Kini}: Kini is an indie game developer through his studio Finite Reflection, and an associate producer at Cartoon Network Games.
  \item \textbf{Gillian Smith}: Smith is an Assistant Professor at WPI. She focuses on game design, AI, craft, and generative design.
  \item \textbf{Kelly Snyder}: Snyder is an Art Producer at Bethesda and previously a Technical Producer at Bungie.
\end{itemize}

\begin{table}[tb]
\begin{center}
\begin{tabular}{ |p{1cm}|c|c|p{4.2cm}| } 
 \hline
 set & labels & total & top three \\ 
 \hline
 Le Doux & 47 & 259 & platform (29), jump (22), pipe-flower (22)\\ 
 \hline
 Del Rosario & 26 & 38 & concept introduction (3), objective (3), completionist reward (2)\\ 
 \hline
 Smith & 21 & 95 & enemy pair (17), staircase (11), ditch (9)\\ 
  \hline
 Smith-Naive & 25 & 118 & multi level (18), enemy pair (17), staircase (13)\\ 
  \hline
Kini & 23 & 28 & hazard introduction (3), hidden powerup (2), pipe trap (2)\\ 
 \hline
Snyder & 34 & 155 & walking enemy (18), power up block (17), coins (13)\\ 
 \hline
\end{tabular}
\caption{A table comparing the characteristics of our six sets of design pattern labels.}
\end{center}
\label{tab:labels}
\end{table}

All five of these experts were asked to label their choice of three levels with labels that established ``a common language/vocabulary that you'd use if you were designing levels like this with another human". Of these experts only Smith had any knowledge of the underlying system. She produced two sets of design patterns for the levels she labeled, one including only those patterns she felt the system could understand and the second including all patterns that matched the above criteria. We refer to these labels as Smith and Smith-Naive through the rest of this section, respectively. 

These experts labeled static images of non-boss and non-underwater Super Mario Bros. levels present in the Video Game Level Corpus (VGLC) \cite{summerville2016vglc}. The experts labeled these images by drawing a rectangle over the level structure in which the design pattern occurred with some string to define the pattern. These rectangles could be of arbitrary size, but we translated each into either a single training example centered on the eight by eight chunk our system requires, or multiple training examples if it was larger than eight by eight.

\begin{figure}[tb]
  \centering
  \includegraphics[width=3in]{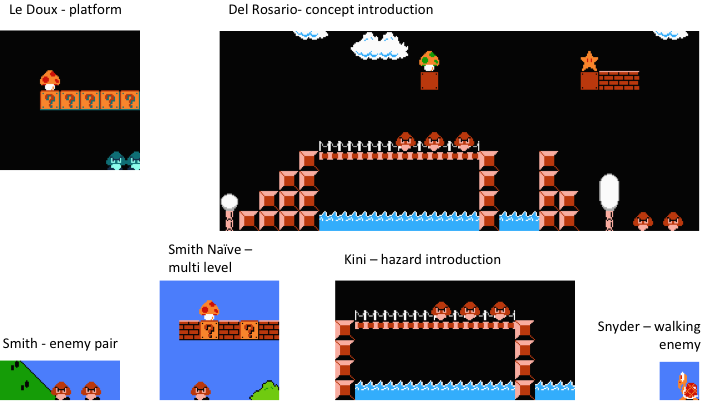}
  \caption{The first example of the the top label in each set of design pattern labels.}
  \label{fig:exampleLabels}
\end{figure}

\begin{table}[tb]
\begin{center}
\begin{tabular}{ |p{1cm}|c|c|c| } 
 \hline
 set & approach & train & test \\ 
 \hline
 Le Doux & RF & 84.06\rpm0.76 & 28.73\rpm1.89\\ 
 \hline
 Le Doux & CNN & 49.06\rpm4.43 & 26.73\rpm3.70\\ 
 \Xhline{4\arrayrulewidth}
 Del Rosario & RF & 86.71\rpm0.78 & 0.77\rpm1.07\\ 
 \hline
 Del Rosario & CNN & 36.08\rpm19.01 & 1.04\rpm0.82\\ 
 \Xhline{4\arrayrulewidth}
 Smith & RF & 92.11\rpm0.89 & 33.28\rpm3.48\\ 
  \hline
 Smith & CNN & 59.76\rpm8.16 & 26.19\rpm5.02\\ 
  \Xhline{4\arrayrulewidth}
 Smith-Naive & RF & 88.49\rpm0.92 & 42.31\rpm2.96\\ 
  \hline
Smith-Naive & CNN & 65.50\rpm19.68 & 36.82\rpm11.84\\ 
  \Xhline{4\arrayrulewidth}
Kini & RF & 90.19\rpm0.69 & 41.93\rpm2.29\\ 
 \hline
Kini & CNN & 44.32\rpm2.89 & 29.07\rpm2.13\\ 
 \Xhline{4\arrayrulewidth}
Snyder & RF & 86.72\rpm1.62 & 1.85\rpm5.56\\ 
 \hline
Snyder & CNN & 49.40\rpm0.77 & 0.00\rpm0.00\\ 
 \hline
\end{tabular}
\caption{A table comparing the accuracy of our random forest label classifier compared to a CNN baseline.}
\end{center}
\label{tab:autoResults}
\end{table}

We include some summarizing information about these six sets of design pattern labels in Table 1. Specifically, we include the total number of labels and the top three labels, sorted by frequency and alphabetically, of each set. Each expert produced very distinct labels, with less than one percent of labels shared between different experts. We include the first example for the top label for each set of design patterns in Figure \ref{fig:exampleLabels}. Even in the case of Kini and Del Rosario, where there is a similar area and design pattern label, the focus differs. We train six separate models, one for each set of design pattern labels (Smith has two). 

\subsection{Label Classifier Evaluation}

In this section we seek to understand how well our random forest classifier is able to identify design patterns in level structure. For the purposes of this evaluation we made use of AlexNet as a baseline \cite{szegedy2016rethinking}, given that a convolutional neural network would be the naive way one might anticipate solving this problem. We chose AlexNet given its popularity and success at similar image recognition tasks. In all instances we trained the AlexNet until its error converged. We make use of a three-fold cross validation on the labels for this and the remaining evaluations. We make use of a three-fold validation to address the variance across even a single expert's labels and due to the small set of labels available for some experts. 

\begin{table}[tb]
\begin{center}
\begin{tabular}{ |p{1cm}|c|c|c| } 
 \hline
 set & No labels & No Auto Tag & Full \\ 
 \hline
 Le Doux & 12.3\rpm6.3 & 152.8\rpm4.8 & \textbf{10.6\rpm5.6}\\ 
 \hline 
 Del Rosario & 10.4\rpm5.0 & 135.7\rpm3.8 & \textbf{9.0\rpm4.4}\\ 
 \hline
 Smith & 11.5\rpm6.1 & 157.2\rpm3.5 & \textbf{10.4\rpm5.6}\\ 
  \hline
 Smith-Naive & 12.7\rpm4.8 & 167.6\rpm4.0 & \textbf{11.5\rpm4.4}\\ 
  \hline
Kini & \textbf{9.4\rpm4.4} & 129.6\rpm3.6 & 10.6\rpm3.3\\ 
 \hline
Snyder & 28.6\rpm9.9 & 144.2\rpm5.0 & \textbf{15.0\rpm9.4}\\ 
 \hline
\end{tabular}
\caption{A table comparing the error in terms of incorrect sprites for our three generators. Smaller values represent fewer mistakes.}
\end{center}
\label{tab:generatorResults}
\end{table}

Our major focus is training and test accuracy across the folds. We summarize the results of this evaluation in Table 2, giving the average training and test accuracies across all folds along with the standard deviation. We note that in all instances our random forest (RF) approach outperformed AlexNet CNN in terms of training accuracy, and nearly always in terms of test accuracy. We note that given more training time AlexNet's training accuracy might improve, but at the cost of test accuracy. We further note that AlexNet was on average one and a half times slower than the random forest in terms of training time. These results indicate that our random forest produces a more general classifier compared to AlexNet.

We note that our random forest performed fairly consistently in terms of training accuracy, at around 85\%, but that the test accuracy varied significantly. Notably, the test accuracy did not vary according to the the number of training samples or number of labels per expert. This indicates that individual experts identify patterns that are more or less easy to classify automatically. Further we note that Snyder and Del Rosario had very low testing error across the board, which indicates a large amount of variance between tagged examples. Despite this, we demonstrate the utility of this approach in the next section.

\subsection{Autoencoder Structure Evaluation}

We hypothesize that the inclusion of design pattern labels into our autoencoder network will improve its overall representative quality. Further, that the use of an automatic label classifier will allow us to gather sufficient training data to train the autoencoder. This evaluation addresses both these hypotheses. We draw upon the same dataset and the same three folds from the prior evaluation and create three variations of our system. The first autoencoder variation has no design pattern labels and is trained on all $8\times 8$ chunks of level instead of only those chunks labeled or autolabeled with a design pattern. Given that this means fewer features and smaller input and output tensors, this model should outperform our full model unless the design pattern labels improve overall representative quality. The second autoencoder variation does not make use of the automatic design pattern label classifier, thus greatly reducing the training data. The last variation is simply our full system. For all approaches we trained till training error converged. We note that we trained a single 'no labels' variation and tested it on each expert, but trained models for the no automatic classifier and full versions of our approach for each expert.

Given these three variations, we chose to measure the difference in structure when the autoencoder was fed the test portions of each of the three folds. Specifically we capture the number of incorrect structure features predicted. This can be understood as a stand in for representation quality, given that the output of the autoencoder for the test sample will be the closest thing the autoencoder can represent to the test sample. 

We give the average number and standard deviation of incorrect structural features/tiles over all three folds in Table 2. We note that the minimum value here would be 0 errors and the maximum value would be $8\times 8\times 30$ or 1920 incorrect structural feature values. For every expert except for Kini, who authored the smallest number of labels, our full system outperformed both variations. While some of these numbers are fairly close between the full and no labels variation, the values in bold were significantly lower according to the paired Wilcoxon Mann Whitney U test ($p<0.001$).

Given the results in Table 3. We argue that both our hypotheses were shown to be correct, granted that the expert gives sufficient labels, with the cut-off appearing to be between Kini's 28 and Del Rosario's 38. Specifically the representation quality is improved when labels are used, and the label classifier improves performance over not applying the label classifier. 

\begin{table*}[tb]
\begin{center}
\begin{tabular}{ |l|c|c|c|c|c|c|c| } 
 \hline
 & Le Doux & Del Rosario & Smith & Smith-Naive & Kini & Snyder \\ 
 \hline
 No labels & 12.3\rpm16.3 & 10.4\rpm5.0 & 11.5\rpm6.1 & 12.7\rpm4.8 & \textbf{9.3\rpm4.4} & 28.6\rpm9.9 \\
 \hline
 Transfer No Auto Tag & 11.1\rpm5.8 & 10.1\rpm5.0 & 11.2\rpm6.1 & 12.6\rpm4.8 & 9.3\rpm4.4 & 16.7\rpm9.8 \\
 \hline
Transfer w/ Auto & 10.8\rpm5.8 & 9.8\rpm5.0 & 11.0\rpm6.0 & 11.8\rpm6.3 & 10.3\rpm4.2 & 16.1\rpm9.6 \\ 
 \hline
 Full & \textbf{10.6\rpm5.6} & \textbf{9.0\rpm4.4} & \textbf{10.4\rpm5.6} & \textbf{11.5\rpm4.4} & 10.6\rpm3.3 & \textbf{15.0\rpm9.4} \\ 
 \hline
\end{tabular}
\caption{A table comparing the transfer learning approach structure errors to the full system structure errors.}
\end{center}
\label{tab:transferResults}
\end{table*}

\subsection{Transfer Evaluation}

A major concern for any co-creative tool based on Machine Learning is training time. In the prior autoencoder evaluation, both the no labels and full versions of our system took hours to train to convergence. This represents a major weakness, given that in some co-creative contexts designers may not want to wait for an offline training process, especially when we anticipate authors wanting to rapidly update their set of labels. Given these concerns, we evaluate a variation of our approach utilizing transfer learning. This drastically speeds up training time by adapting the weights of a pretrained network on one task to a new task.

We make use of student-teacher or born again neural networks, a transfer learning approach in which the weights of a pre-trained neural network are copied into another network of a different size. In this case we take the weights from our no labels autoencoder from the prior evaluation, copy them into our full architecture, and train from there. We construct two variations of this approach, once again depending on the use of the random forest label classifier or not. We compare both variations to the full and no labels system from the prior evaluation, using the same metric.

We present the results of this evaluation in Table 4. We note that, while the best performing variation did not change from the prior variation, in all cases except for the Kini models, the transfer approaches got closer to the full variation approach, sometimes off by as little as a fraction of one structure feature. Further, these approaches were significantly faster to train, with the no automatic labeling transfer approach training in an average of 4.48 seconds and the automatic labeler transfer approach training in an average of 144.92 seconds, compared to the average of roughly five hours of the full approach on the same computer. This points to a clear breakdown in when it makes sense to apply what variation of our approach, depending on time requirements and processing power. In addition, it continues to support our hypotheses concerning the use of automatic labeler and personal level design pattern labels.

\begin{figure}[tb]
  \centering
  \includegraphics[width=3in]{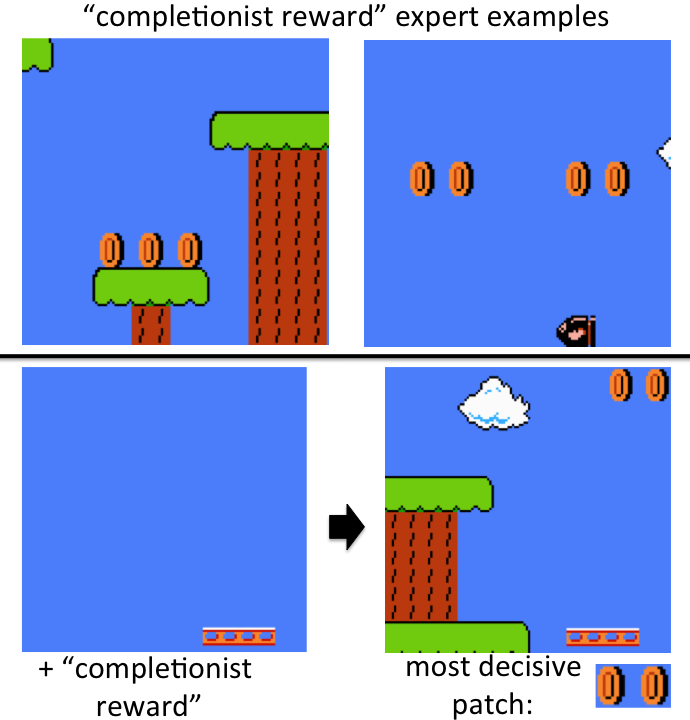}
  \caption{Above: the two examples of the pattern ``completionist reward'' labeled by the expert Dee Del Rosario. Below: example of the system given the input on the left and that label and its output.}
  \label{fig:example}
\end{figure}

\subsection{Qualitative Example}
We do not present a front-end or interaction paradigm for the use of this Explainable PCGML system, as we feel such implementation details will depend upon the intended audience. 
However, it is illustrative to give an example of how the system could be used. 
In Figure \ref{fig:example} we present an example of the two training examples of the pattern ``completionist reward'' labeled by the expert Dee Del Rosario. 
The full system, including the random forest classifier, trains on these examples (and the other labels from Del Rosario), and  is then given as input the eight by eight chunk with only the floating bar within it on the left of the image along with the desired label ``completionist reward''. One can imagine that Del Rosario as a user wants to add a reward to this section, but doesn't have any strong ideas. Given this input the system outputs the image on the right. 

We asked Del Rosario what they thought of the performance of the system and whether they considered this output matched their definition of completionist reward. They replied ``Yes -- I think? I would because I'm focusing on the position of the coins.'' We note that Del Rosario did not see the most decisive patch when making this statement, which we extracted as in \cite{olah2018building}. This clearly demonstrates some harmony between the learned model and the design intent. However, Del Rosario went on to say ``I think if I were to go... more strict with the definition/phrase, I'd think of some other configuration that would make you think, `oooooh, what a tricky design!!' ''. This indicates a desire to further clarify the model. Thus, we imagine an iterative model is necessary for a tool utilizing this system and a user to reach a state of harmonious interaction.

\section{Conclusions}

In this paper, we present an approach to explainable PCGML (XPCGML) through user-authored design pattern labels over existing level structure.
We evaluate our autoencoder and random forest labeler components on levels labeled by game design experts. 
These labels serve as a shared language between the user and level design agent, which allows for the possibility of explainability and meaningful collaborative interaction. 
We intend to take our system and incorporate it into a co-creative tool for novice and expert level designers. To the best of our knowledge this represents the first approach to explainable PCGML.

\section{Acknowledgements}

We want to thank our five expert volunteers. This material is based upon work supported by the National Science Foundation under Grant No. IIS-1525967. We would also like to thank the organizers and attendees of Dagstuhl Seminar 17471 on Artificial and Computational Intelligence in Games: AI-Driven Game Design, where the discussion that lead to this research began.

\bibliographystyle{aaai}
\bibliography{aaai}

\end{document}